%% file: main.tex
\documentclass[10pt,twocolumn,letterpaper]{article}
\usepackage[algorithms]{wacv}
\usepackage{graphicx}
\usepackage{amsmath}
\usepackage{amssymb}
\usepackage{booktabs}
\usepackage{xcolor}
\usepackage{array}
\usepackage{multirow}
\usepackage{color}
\usepackage{colortbl}
\usepackage{framed}
\usepackage{bm}
\usepackage{bbm}
\usepackage{xspace}
\usepackage{enumitem}
\usepackage{algorithm}
\usepackage{listings}
\usepackage{url}
\usepackage{pifont} 
\usepackage{caption}
\usepackage{ifthen}
\usepackage{comment}
\definecolor{Gray}{gray}{0.9}
\definecolor{LightCyan}{rgb}{0.88,0.95,1}
\definecolor{blond}{rgb}{0.98, 0.94, 0.75}
\definecolor{green}{rgb}{0.0, 0.62, 0.42}
\definecolor{lightgray}{rgb}{0.83, 0.83, 0.83}
\usepackage[pagebackref,breaklinks,colorlinks]{hyperref}
\usepackage[accsupp]{axessibility} 

\usepackage[capitalize]{cleveref}
\crefname{section}{Sec.}{Secs.}
\Crefname{section}{Section}{Sections}
\Crefname{table}{Table}{Tables}
\crefname{table}{Tab.}{Tabs.}

\newcommand{\tformer}{T-Former\xspace}
\newcommand{\ours}{PQR\xspace}
\newcommand{\framework}{PQR\xspace}
\DeclareMathOperator*{\argmax}{arg\,max}
\newcommand{\cmark}{\ding{51}}%
\newcommand{\xmark}{\ding{55}}%
\newcommand{\tit}[1]{\smallbreak\noindent\textbf{#1.}}

\newcommand\blfootnote[1]{%
  \begingroup
  \renewcommand\thefootnote{}\footnote{#1}%
  \addtocounter{footnote}{-1}%
  \endgroup
}

\begin{document}

\title{Perceive, Query \& Reason: Enhancing Video QA with \\ Question-Guided Temporal Queries} 

\author{
Roberto Amoroso$^{3,*}$ \quad Gengyuan Zhang$^{1,2,*}$ \quad Rajat Koner$^1$ \\ Lorenzo Baraldi$^3$ \quad Rita Cucchiara$^{3}$ \quad Volker Tresp$^{1,2}$ \\
$^1$LMU Munich \quad $^2$MCML \quad $^3$University of Modena and Reggio Emilia \\
{\tt\small roberto.amoroso@unimore.it \quad zhang@dbs.ifi.lmu.de}
}
\maketitle

\input{sections/0-Abstract}
\input{sections/1-Introduction}
\input{sections/2-Related}

\input{sections/3-Method}
\input{sections/4-Experiments}
\input{sections/5-Conclusion}

{\small
\bibliographystyle{ieee_fullname}
\bibliography{main}
}

\clearpage
\appendix
\input{sections/X-Suppl}

\end{document}

%% file: sections/0-Abstract.tex
\begin{abstract}
Video Question Answering (Video QA) is a challenging video understanding task that requires models to comprehend entire videos, identify the most relevant information based on contextual cues from a given question, and reason accurately to provide answers. Recent advancements in Multimodal Large Language Models (MLLMs) have transformed video QA by leveraging their exceptional commonsense reasoning capabilities. This progress is largely driven by the effective alignment between visual data and the language space of MLLMs. However, for video QA, an additional space-time alignment poses a considerable challenge for extracting question-relevant information across frames. In this work, we investigate diverse temporal modeling techniques to integrate with MLLMs, aiming to achieve question-guided temporal modeling that leverages pre-trained visual and textual alignment in MLLMs. We propose \tformer, a novel temporal modeling method that creates a question-guided temporal bridge between frame-wise visual perception and the reasoning capabilities of LLMs. Our evaluation across multiple video QA benchmarks demonstrates that \tformer competes favorably with existing temporal modeling approaches and aligns with recent advancements in video QA.
\blfootnote{$^*$Equal contribution.}
\end{abstract}

%% file: sections/1-Introduction.tex
\section{Introduction}
\label{sec:intro}
In recent years, video understanding tasks have gained increasing attention, driven by advancements in multimodal learning and vision-language models~\cite{caffagni2024r,xu2023multimodal,damonlpsg2023videollama,liu2024llavanext}. Videos are ubiquitous on the Internet and their inherently dynamic and complex nature poses a significant challenge for video understanding~\cite{tang2023video,zhang2023multi,Maaz2023VideoChatGPT,zhong2022video}.

Video Question Answering (Video QA) emerges as a particularly fundamental and challenging task within this domain, requiring models to not only perceive videos for recognition-level tasks like classification or segmentation but also to comprehend and rationalize the video content in response to specific questions. 
The challenge is twofold: models must process the temporal and dynamic nature of videos with their sparse and heterogeneous event distributions, while simultaneously developing a deep understanding that connects these events through commonsense and temporal reasoning. This necessitates a blend of model abilities including visual perception, question contextualization, and answer reasoning.

\begin{figure}
    \centering
    \includegraphics[width=1.\linewidth]{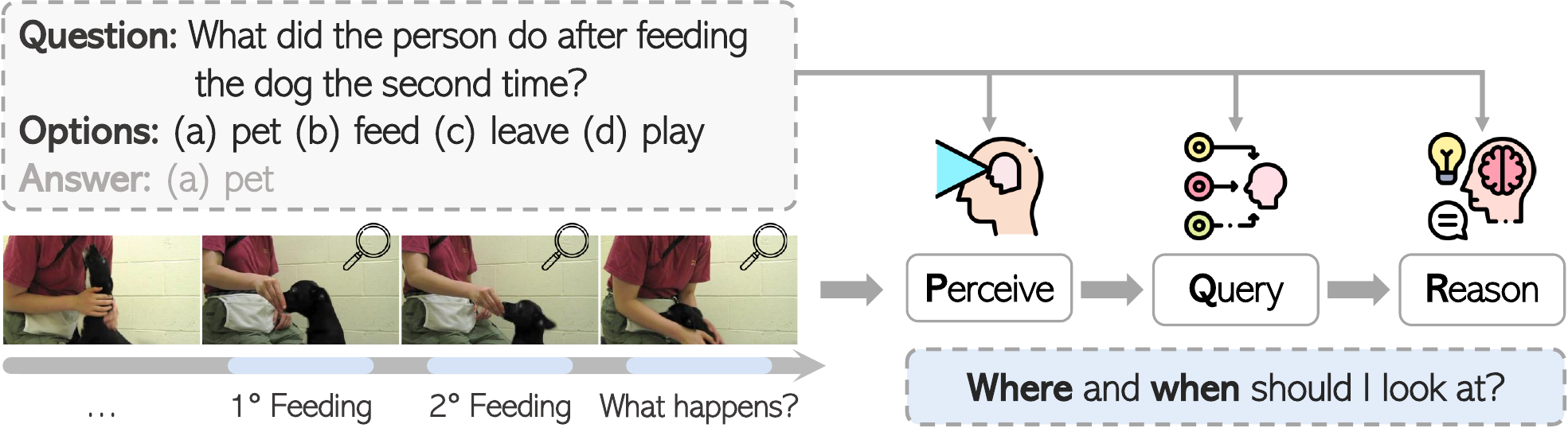}
    \vspace{-4mm}
    \caption{
    Adapting LLMs' visual reasoning capabilities to video QA requires extracting the most relevant video features based on the input question. Our approach addresses this challenge by extracting question-guided temporal features from dynamic frame sequences, enabling accurate and context-aware reasoning.}
    \label{fig:teaser}
\end{figure}

The advent of Large Language Models (LLMs) and Multimodal Large Language Models (MLLMs) has marked significant progress in visual reasoning, including image QA~\cite{li2023blip,chen2022pali,liu2023llava} and video QA~\cite{damonlpsg2023videollama,yu2023self} tasks.
MLLMs exhibit strong general knowledge and reasoning capabilities, excelling in interpreting visual content for video QA applications.
A common paradigm in MLLMs involves integrating a visual encoder for embedding video frames and a projection network that compresses visual features and aligns them with LLMs. Conventional compression strategies, such as sparse sampling and pooling-based methods~\cite{luo2021clip4clip,zhou2017adaptive,lei2022revealing,huang2021tada}, face limitations when dealing with lengthy and dynamic videos. Recent studies~\cite{damonlpsg2023videollama,zhu2023minigpt,zhao2023mmicl,alayrac2022flamingo} have achieved more promising results by adopting attention-based models like Q-Former~\cite{li2023blip,dai2023instructblip}. The Q-Former extracts prominent visual features from the visual encoder into a small set of learnable queries, which are then aligned with the textual embeddings from a pre-trained LLM via a linear layer.
This paradigm has proven effective across various vision-language tasks~\cite{zhu2023minigpt,damonlpsg2023videollama,yu2024connecting,zhao2023mmicl}, demonstrating that visual features can be compressed without compromising the overall representation while also reducing computational costs.

However, directly adapting image-based Q-Former approaches to video data often results in suboptimal performance due to the spatio-temporal complexities and varying information density inherent in videos~\cite{yu2023self}. 
We hypothesize that pre-trained image-level models lack the necessary spatio-temporal inductive bias for video understanding. This limitation hinders the seamless integration of LLMs with video content, especially in scenarios involving complex dynamics and diverse information distributions.

In this work, we introduce \tformer, a question-guided temporal querying Transformer designed to efficiently sample and learn video-specific features tailored to a given question. Unlike existing methods, \tformer leverages knowledge from image-based pre-training without requiring additional video pre-training. With \tformer, we propose a new LLM-based model, \ours, based on three core principles: \textit{\textbf{P}erceive} to extract framewise question-guided spatial features from input videos, \textit{\textbf{Q}uery} to extract a fixed-size set of question-guided temporal visual features from the whole sequence, and \textit{\textbf{R}eason} to integrate these extracted features with LLMs for generating answers, as illustrated in Fig.~\ref{fig:teaser}.
\tformer employs an efficient temporal sampling strategy that incorporates an inductive bias for video queries by initializing learnable input queries with discrete key video event features. This design bridges the gap between pre-trained image-level knowledge and video-specific representations while focusing on relevant question-guided information. By doing so, \tformer ensures a seamless transition from generic to question-tailored video features.

Our evaluations demonstrate that \tformer outperforms alternative temporal modeling techniques, achieving peak performance at a sampling ratio equivalent to four frames, thereby effectively harnessing the potential of LLMs. Notably, \ours sets a new state-of-the-art benchmark in video QA, surpassing existing methods.

To sum up, the contributions of this paper are as follows:
\begin{itemize}[noitemsep,topsep=0pt]
 \item We propose \tformer as a simple yet effective temporal modeling method that can effectively capture question-guided information from video frames;
 \item We show that \tformer achieves superior performance on video QA while maintaining a good balance of computation complexity;
 \item We present \ours, a novel LLM-based model achieving competitive performance across multiple video QA benchmarks, thereby redefining the state-of-the-art in this field.
\end{itemize}

%% file: sections/2-Related.tex
\section{Related Works}

\subsection{Video Question Answering}
Video Question Answering is a fundamental task of video-language understanding~\cite{zhong2022video}.
Early approaches to video language modeling primarily relied on sequential neural networks, such as Long Short-Term Memory networks~\cite{le2020neural}, 3D Convolutional Neural Networks~\cite{li2019beyond}, and Graph Neural Networks~\cite{xiao2022video,peng2021progressive}. These models were effective in capturing temporal and spatial information but faced limitations in handling complex video data and contextual reasoning.

With the advent of Transformer-based visual encoders and novel paradigms for video-language pre-training~\cite{radford2021learning,wang2022internvideo,wang2023all,lei2022revealing, bain2021frozen, yang2022zero,yang2021just, buch2022revisiting,yu2023self}, encoder-decoder models have achieved significant advancements in the video QA task. These models benefit from the Transformers' ability to capture long-range dependencies and contextual information, leading to improved performance in understanding and reasoning about video content.
The development of video QA has been driven by diverse benchmark datasets spanning multiple domains, including movies and TV shows~\cite{lei2018tvqa, lei2020more, xu2017video, tapaswi2016movieqa}, as well as wild, online videos~\cite{xiao2021next, wu2021star, xu2017video, yu2019activitynet, grunde2021agqa}. Notably, datasets like NeXT-QA~\cite{xiao2021next} and STAR~\cite{wu2021star} have introduced more challenging aspects of video understanding, emphasizing temporal reasoning and causal relationships, which align closely with our work's focus on question-guided temporal modeling.

\begin{figure*}[!t]
    \centering
    \includegraphics[width=\textwidth]{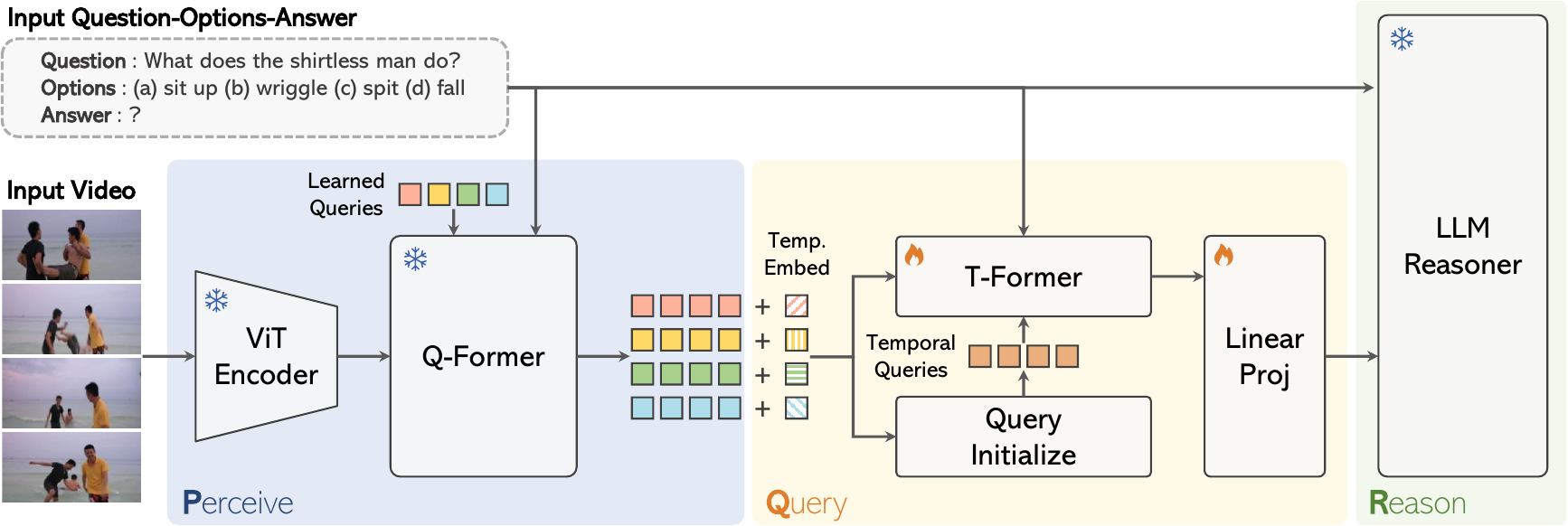}
    \caption{
    Our model consists of three primary stages: \textit{\textbf{P}erceive}, \textit{\textbf{Q}uery}, and \textit{\textbf{R}eason}. 
    The \textit{Perceive} stage employs a ViT encoder and a Q-Former to extract spatial visual features from each frame independently, conditioned on the input question.
    The \textit{Query} stage introduces \tformer, our question-guided temporal querying Transformer that captures the most relevant temporal information conditioned on both the question and the visual context. 
    Finally, in the \textit{Reason} stage, the condensed visual-temporal features are integrated with the question and answer options before being fed into a frozen LLM as a reasoning agent to rationalize and answer the question.
    }
    \label{fig:framework}
\end{figure*}

\subsection{Temporal Modeling for Video QA}
Temporal feature extraction plays a crucial role in video understanding due to the dense, dynamic, and redundant content of video data.
\cite{buch2022revisiting, lei2022revealing, lei2021less} propose sparse sampling based on dataset biases. While these methods can be effective, they often struggle with complex videos containing multiple events or subtle temporal dependencies, limiting their applicability in real-world scenarios.
Several works also adopt a ``first temporal, then spatial'' strategy to enhance video understanding by training a temporal selection module. TranSTR~\cite{li2023discovering} uses Adaptive Temporal Rationalization to select critical frames, whereas  MIST~\cite{gao2023mist} leverages a temporal attention layer to aggregate frame-wise features. These approaches demonstrate the importance of selective temporal processing but often involve complex architectural designs.

Recent advances in LLMs have opened new possibilities for video QA. However, integrating video inputs with LLMs remains challenging, particularly due to the quadratic computational complexity associated with processing long input sequences.
In this work, we propose \tformer, a question-guided temporal querying Transformer that, 
achieves competitive performance while mitigating the computational overhead.

\subsection{Video QA with LLMs}
Pre-trained LLMs have drawn significant attention in recent years owing to their emergent capabilities of instruction following~\cite{zheng2023judging, chung2022scaling, alpaca} and human-like commonsense reasoning~\cite{kojima2022large, liu2023evaluating,yao2022react}.
In the field of visual understanding, Multimodal Large Language Models (MLLMs) attempt to merge the capabilities of LLMs with visual data to address complex vision tasks, such as video QA~\cite{li2023blip}. 
These models have demonstrated superior visual reasoning abilities compared to traditional models, establishing a new baseline in visual comprehension.
The application of LLMs to video QA presents both promising opportunities and unique challenges. The inherently multi-event nature of video content introduces complexities not present in static images. Videos require models to navigate and reason across temporal sequences, a task that demands more than just the integration of visual and textual modalities.
To address these challenges, the Q-Former emerges as a powerful component to compress spatial visual features~\cite{damonlpsg2023videollama,dai2023instructblip,li2023blip}, thus facilitating the alignment of these features with the reasoning capabilities of LLMs.
By adopting and further developing the Q-Former architecture, our work follows this direction of utilizing an LLM as a powerful agent for reasoning on video data and explores how to extract temporal visual features from videos and align them with MLLMs.

%% file: sections/3-Method.tex
\section{Proposed Method}

In this section, we introduce \ours, our LLM-based video QA model, alongside \tformer, a novel question-guided temporal querying Transformer.
Our model, illustrated in \cref{fig:framework}, consists of three main modules: (1) \textit{\textbf{P}erceive}: framewise visual encoding of input videos conditioned on given questions; (2) \textit{\textbf{Q}uery}: \tformer for selecting question-relevant temporal information across frames; and (3) \textit{\textbf{R}eason}: an adapted LLM as a reasoning agent for question answering. 

We start by formulating the video QA task. Given a video $v$ as a sequence of $n$ image frames $I^{1:n}$, where $I^i$ represents the $i$-th frame of the video, and a question $q$, a model $f(\cdot)$ assigns a probability to an answer $a$ from the answer space $\mathcal{A}$. This can be formulated as:
\begin{equation}
    \hat{a} = \argmax_{a\in \mathcal{A}} f(a | v, q, \mathcal{A}),
\end{equation}
where $\hat{a}$ is the predicted answer given the video and questions.
The answer space $\mathcal{A}$ can be large for open-end QA or preset for multiple-choice QA.

\subsection{Perceive: Visual Encoding}
State-of-the-art visual encoders such as CLIP~\cite{radford2021learning} and BLIP-2~\cite{li2023blip} have demonstrated remarkable zero-shot capabilities in generating visual features, making it a common practice~\cite{yu2023self} to leverage these powerful features for video QA tasks.
Despite their success, these models are restricted to extracting vision-only features that ignore other modalities, like textual questions in our task. 
This leads to question-agnostic visual feature extraction and a lack of contextual relevance to complex questions.

To address this limitation, our \ours employs a question-guided visual feature extraction mechanism.
Specifically, we adopt the instruction-aware Q-Former proposed by~\cite{dai2023instructblip} to extract question-guided features from each frame.
In more detail, given a sequence $I^{1:n}$, we encode each frame \textit{independently} to get a sequence $\mathbf e_f^{1:n}$ of framewise visual tokens:
\vspace{-4mm}
\begin{equation}
    \mathbf{e}_f^{i} = \operatorname{VisualEncoder}(I^{i}, q, \mathcal{A}), \forall i \in \{1, 2, \ldots, n\},
\end{equation}
where $\mathbf{e}_f^i$ represents the visual features of $i$-th frame, and $\operatorname{VisualEncoder}$ refers to both the frame encoder and Q-Former for simplicity. This step aims to \textit{perceive} and extract spatial visual information from raw frames.

\subsection{Query: \tformer}

\begin{figure}[tbp]
    \centering
    \includegraphics[width=\linewidth]{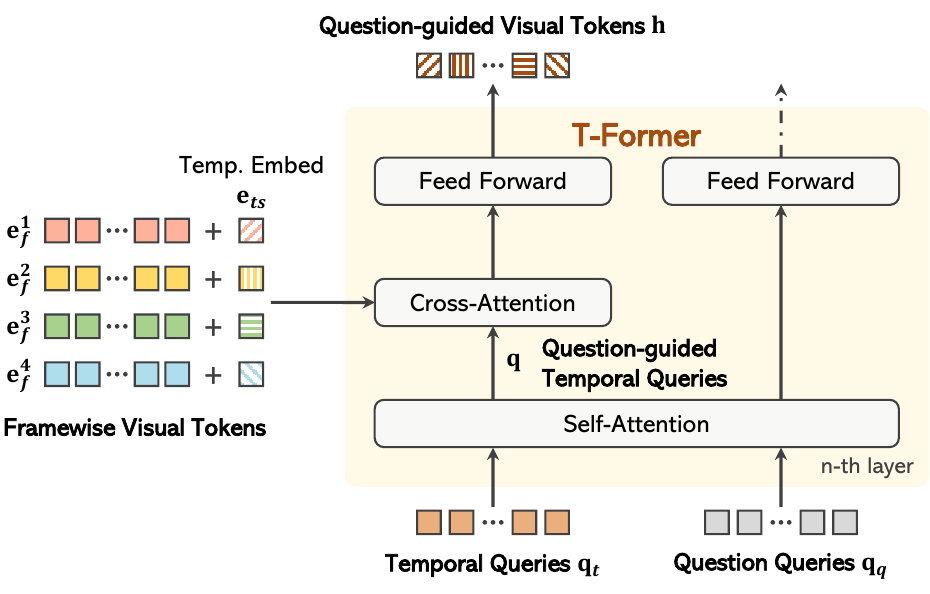}
    \caption{Overview of \tformer, a Temporal Querying Transformer. It adopts 1) a self-attention layer between temporal queries and question queries and 2) cross-attention layers between query tokens and the full-length sequence of visual tokens. 
    }
    \label{fig:model}
\end{figure}

Handling a full sequence of framewise visual tokens $\mathbf{e}_f^{i:n}$ leads to temporal information redundancy across frames and is also computationally demanding due to the complexity of LLMs.
To mitigate these challenges, we propose \tformer, a question-guided \textbf{\textit{t}}emporal querying trans\textbf{\textit{former}}, designed to select the most relevant visual features across time based on the input question, as presented in \cref{fig:model}.

\tformer integrates three distinct types of input tokens: 1) the concatenation of framewise visual tokens $\mathbf e_f^{1:n}$ from the \textit{perceive} step, 2) temporal queries $\mathbf{q}_t$ of a fixed length to \textit{query} temporally relevant information from the whole framewise visual token sequence, and 3) question queries $\mathbf{q}_q$ to condition the output tokens on the questions. \tformer first contextualizes query tokens with self-attention between $\mathbf{q}_t$ and  $\mathbf{q}_q$; and then performs cross-attention between $\mathbf{q}_t$ and $\mathbf{e}_f$ to select question-relevant visual information over time.
We elucidate each module as follows.

\tit{Temporal Query Initialization} The core of our approach lies in the initialization of a small number of fixed-length temporal queries, $\mathbf{q}_t$, which are specific to the input video and the question. Existing methods~\cite{damonlpsg2023videollama,li2023blip,dai2023instructblip,zhu2023minigpt} typically employ learnable queries with random initialization to compress features. However, these randomly initialized static queries require extensive video-based pre-training, as they lack any inherent alignment with the spatio-temporal dynamics of video events. Moreover, even after pre-training, these static queries remain identical across all videos, failing to capture video-specific contextual information. 
To overcome these limitations, we propose a novel strategy that replaces static, pre-trained queries with dynamic queries sampled directly from the input video features. By grounding the query tokens in the unique characteristics of the input video, our approach introduces a more effective inductive bias for \tformer that aligns with Q-Former's image understanding capabilities. This dynamic grounding not only enables a deeper understanding of the video but also facilitates more precise responses to video-related questions. 

We investigate four distinct sampling strategies for temporal query initialization: uniform sampling, random sampling, $k$-means sampling, and $k$-medoids sampling:
\begin{equation}
\mathbf{q}_t = \operatorname{sampler}(\mathbf{e}_f^{1}, \mathbf{e}_f^{2}, ..., \mathbf{e}_f^{n})
\end{equation}

The size of temporal queries $\mathbf{q}_t$ is considered a crucial model hyperparameter, governing two essential aspects. Firstly, it defines the sampling ratio of \tformer, thereby controlling how much the input sequence is compressed. Secondly, it influences the length of visual tokens fed into LLMs. Hence, a higher number of temporal queries increases computational overheads within LLMs. In \Cref{sec:design}, we investigate the effects of various sampling methods and sampling ratios.

\tit{Question-guided Temporal Query Tokens}
To condition the temporal queries $\mathbf{q}_t$ based on the questions, we concatenate $\mathbf{q}_t$ with the question queries $\mathbf{q}_q$. These question queries are generated through the tokenization process applied to the question $q$ and the answer options $\mathcal{A}$. To facilitate interaction between different modalities, we employ self-attention layers to contextualize both $\mathbf{q}_t$ and $\mathbf{q}_q$. This process is formulated as follows:
\begin{equation}
\mathbf{Q}= \mathbf{K}= \mathbf{V} = \phi(\operatorname{concat}(\mathbf{q}_{t}, \mathbf{q}_q)),
\end{equation}
\vspace{-5mm}
\begin{equation}
\mathbf{q} = \operatorname{self-attention}(\mathbf{Q}, \mathbf{K}, \mathbf{V}),
\end{equation}

where $\phi$ represents the linear projection function of the attention~\cite{vaswani2017attention} mechanism and $\mathbf{q}$ denotes the resulting question-guided temporal query tokens.

\tit{Query-Frame Cross-attention}
The sampled temporal queries still lack spatio-temporal information from videos. Therefore, revisiting the video to enhance the queries with more detailed and fine-grained video-specific information becomes imperative.

To this aim, given the contextualized queries $\mathbf{q}$, we employ cross-attention between $\mathbf{q}$ and the entire sequence of video features. The process also incorporates learnable timestamp embeddings $\mathbf{e}_{ts}$ to capture temporal dependencies, as formulated in the following:
\begin{equation}
\quad \hat{\mathbf{e}}_f^{i} = \mathbf{e}_f^{i} + \mathbf{e}_{ts},
\end{equation}
\vspace{-6mm}
\begin{equation}
\mathbf{Q} = \psi_q(\mathbf{q}), \quad \mathbf{K} = \mathbf{V} = \psi_{kv}(\operatorname{concat}(\hat{\mathbf{e}}_f^{1}, ..., \hat{\mathbf{e}}_f^{n})),
\end{equation}
\vspace{-4mm}
\begin{equation}
\mathbf{h} = \operatorname{cross-attention}(\mathbf{Q}, \mathbf{K}, \mathbf{V}).
\end{equation}

Finally, we pass the final \tformer hidden state $\mathbf{h}$ through a feed-forward layer to derive a representation $\mathbf{t}_v$ that integrates video content, question context, and required information for a precise answer. Importantly, our cross-attention mechanism between the temporal queries and the full-length sequence of video tokens enables the \tformer to perform question-guided spatio-temporal feature extraction. This mechanism enriches temporal query tokens by integrating additional relevant information from the full sequence of video tokens, compensating for the partial context captured during initial sampling and ensuring a comprehensive, question-specific understanding of the video.

\begin{table}
\centering
\setlength{\tabcolsep}{8pt}
\resizebox{0.95\linewidth}{!}{
\begin{tabular}{c cc ccc}
\toprule
\multirow{2}{*}{\shortstack{\\ \textbf{Sampling} \\ \textbf{Strategy}}} &\multirow{2}{*}{\shortstack{\\ \textbf{\#Input} \\ \textbf{Frames}}}  & \multicolumn{4}{c}{\textbf{NExT-QA}} \\
\cmidrule(lr){3-6}
 & & \textbf{Tem.} & \textbf{Cau.} & \textbf{Des.} & \textbf{Avg.} \\
\midrule
        Single Frame  & 1  & 67.7 & 72.3 & 79.0 & 71.9\\
        \midrule
        Concatenate  & 4  & 66.1 & 72.3 & 80.6 & 71.6 \\
        Concatenate  & 16 &  69.6 & 74.0 & 81.1 & 73.2 \\
        \midrule
        Mean-pooling & 4 & 64.5 & 69.6 & 77.2 & 69.1 \\
        Mean-pooling & 16 & 64.6 & 70.5 & 79.6 & 70.0 \\
        \midrule
        Spatio-Temporal  & 4  & 64.5 & 70.7 & 77.5 & 69.8 \\
        Spatio-Temporal  & 16 &  67.7 & 72.6 & 79.5 & 72.1 \\
        \midrule
        \textbf{\tformer (Ours)}  & 16 & \textbf{72.8} & \textbf{76.9} & \textbf{84.7} & \textbf{76.6}\\
\bottomrule
\end{tabular}}
\caption{Comparing \tformer with other temporal modeling methods. \tformer outperforms other sampling methods.}
\label{tab:temporal-baseline}
\end{table}

\subsection{Reason: LLMs as reasoning agents}
The reasoning component of our framework leverages an LLM to reason and extract meaningful insights from the rich visual-temporal representations produced by our \tformer.
Aggregated visual features from \tformer, question tokens, and answer option tokens from the LLM tokenizer are fed as inputs to the LLM reasoner. We incorporate a linear projection layer to reformat the visual features and ensure alignment with the LLM's feature space:
\begin{equation}
\mathbf{h}_{vq\mathcal{A}} = \operatorname{LLM}(\operatorname{linear-proj}(\mathbf{t}_v),q, \mathcal{A}).
\end{equation}

The LLM's hidden state $\mathbf{h}_{vqA}$ represents a unified embedding that captures the complex interactions between the contextualized representations of video, question, and answer space $(v, q, \mathcal{A})$. For multiple-choice video QA tasks, we append a linear projection layer, as a common practice, to generate logits for cross-entropy loss computation.

This enhanced reasoning module enables sophisticated multi-modal understanding, leveraging the pre-trained knowledge of the LLM while maintaining efficiency through our selective temporal feature extraction approach.

%% file: sections/4-Experiments.tex
\begin{table}
\centering
\setlength{\tabcolsep}{8pt}
\resizebox{0.7\linewidth}{!}{
\begin{tabular}{c cccc}
\toprule
\multirow{2}{*}{\shortstack{\\ \textbf{Sampling} \\ \textbf{Strategy}}} & \multicolumn{4}{c}{\textbf{NExT-QA}} \\
\cmidrule(lr){2-5}
 & \textbf{Tem.} & \textbf{Cau.} & \textbf{Des.} & \textbf{Avg.} \\
\midrule
Uniform & 70.8 & 74.9 & 83.8 & 75.0 \\
Random & 71.7 & 75.0 & 84.6 & 75.4 \\
$k$-means & 71.9 & 76.7 & 83.4 & 76.2 \\
$k$-medoids & \textbf{72.8} & \textbf{76.9} & \textbf{84.7} & \textbf{76.7} \\
\bottomrule
\end{tabular}}
\caption{Ablation studies of different sampling methods for temporal query initialization. The number of input frames is 16. The clustering-based methods outperform other techniques.} 
\label{tab:sample}
\end{table}

\section{Experiments}
\subsection{Task and Datasets}
We evaluate \ours in a multiple-choice video QA setting, where each question is accompanied by a fixed number of answer choices. Our primary evaluation is conducted on the NExT-QA dataset~\cite{xiao2021next}, which serves as our development benchmark for exploring various architectural decisions. To ensure comprehensive validation, we extend our investigation to three additional competitive video QA benchmarks: NExT-QA, STAR~\cite{wu2021star}, and How2QA~\cite{li2020hero}, as well as one video event prediction benchmark: VLEP~\cite{lei2020more}.
Notably, \textbf{\ours is trained exclusively on target datasets} without requiring any video-based pre-training phase.

We follow established evaluation protocols~\cite{dai2023instructblip, damonlpsg2023videollama} for each benchmark to ensure comparability with prior works. Further details about the experimental setup are provided in the appendix.

\subsection{Implementation Details}
Our framework adopts the ViT-G/14 from~\cite{fang2023eva} and the instruction-aware Q-Former from~\cite{dai2023instructblip} as the pre-trained visual encoder for frame processing. The reasoning component in \ours is powered by a frozen Vicuna-7B model~\cite{zheng2023judging}, an instruction-tuned LLM built on top of the LLaMA architecture~\cite{touvron2023llama}. 
Within \framework, we integrate two layers of our \tformer, which is based on a BERT-structured Transformer. To ensure a smooth integration between \tformer~’s visual features and the LLM’s feature space, we utilize a linear projection layer. Additionally, a trainable linear layer is appended to the LLM for fine-tuning in the multi-choice video QA setting. During training, both the visual encoder and the LLM remain frozen, and only \tformer and the linear projection layers are trained.

\tit{LLM Prompts}
For all tasks, we adhere to the protocol of~\cite{li2023blip, dai2023instructblip} and adopt a canonical context template: ``\textit{Question: }[\textit{\textless Question\textgreater}]\textit{. Options: }[\textless \textit{Option Lists}\textgreater]\textit{. Answer:}'' used for textual inputs of Q-Former, \tformer, and the LLM reasoner. 
For the future event prediction task on VLEP, we use the same question template, ``\textit{Which event is more likely to happen right after?}'', for all samples.

\tit{Training details}
Our \framework is trained on a single NVIDIA A6000 GPU with 48GB of memory. We employ the AdamW optimizer with a linear warmup and cosine annealing scheduler. Detailed batch sizes and learning rates for each dataset are provided in the appendix.

\begin{table}
\centering
\setlength{\tabcolsep}{8pt}
\resizebox{0.8\linewidth}{!}{
\begin{tabular}{c cccc}
\toprule
\multirow{2}{*}{\shortstack{\\ \textbf{Temporal Queries} \\ \textbf{Initialization}}} & \multicolumn{4}{c}{\textbf{NExT-QA}} \\
\cmidrule(lr){2-5}
 & \textbf{Tem.} & \textbf{Cau.} & \textbf{Des.} & \textbf{Avg.} \\
\midrule
Learnable &  63.0   & 65.2 & 73.2 & 65.7 \\
Sampled &  \textbf{72.8}   & \textbf{76.9} & \textbf{84.7} & \textbf{76.7} \\
\bottomrule
\end{tabular}
}
\vspace{-1mm}
\caption{Effect of using learnable temporal queries instead of sampled temporal queries. We found that sampled temporal queries achieve better performance.}
\label{tab:learnable_ablation}
\end{table}

\begin{table}[tbp]
    \centering
    \centering
    \setlength{\tabcolsep}{8pt}
    \resizebox{0.8\linewidth}{!}{
    \begin{tabular}{c cccc}
    \toprule
     \multirow{2.5}{*}{\textbf{\#Hidden Layers}} & \multicolumn{4}{c}{\textbf{NExT-QA}} \\
    \cmidrule(lr){2-5}
     & \textbf{Tem.} & \textbf{Cau.} & \textbf{Des.} & \textbf{Avg.} \\
    \midrule
    12 &  67.1   & 72.1 & 80.1 & 71.7 \\
    8 &  69.1   & 74.8 & 83.4 & 74.3 \\
    4 &  71.1   & 75.4 & 83.9 & 75.3 \\
    2 &  \textbf{72.8}   & \textbf{76.9} & \textbf{84.7} & \textbf{76.7} \\
    1 &  70.1   & 75.3 & 84.3 & 75.1 \\
    \bottomrule
    \end{tabular}}
    \vspace{-1mm}
    \caption{Effect of the number of hidden layers. Overall, \tformer with two hidden layers has achieved the best performance.} 
    \label{tab:nl}
\end{table}

\subsection{Evaluation Metrics}
We report the standard answer accuracy metric for the multiple-choice video QA tasks. Each dataset has a different number of candidate answer options: NExT-QA provides five options per question, STAR and How2QA provide four options, and VLEP offers two options. Additionally, NExT-QA categorizes questions into three types: temporal, causal, and descriptive, while STAR delineates four types: interaction, sequence, prediction, and feasibility. 

\subsection{Comparison to Other Temporal Modeling}
We compare our \tformer with several temporal modeling baselines on the NExT-QA dataset, including:
\begin{itemize}[topsep=0pt, partopsep=0pt, parsep=0pt, itemsep=0pt]
    \item \textit{Single Frame Sampling}: A single frame is randomly sampled from the whole video sequence.
    \item \textit{Frame Concatenation}: Visual tokens from each frame are concatenated before being fed into the LLM.
    \item \textit{Mean-Pooling}: Framewise outputs from the visual encoder are merged via mean pooling.
    \item \textit{Spatio-Temporal Transformers}: The ViT features from all frames are concatenated in sequence and input into the Q-Former, which jointly extracts spatio-temporal features before passing them to the LLM.
\end{itemize}
Our results, summarized in Tab.~\ref{tab:temporal-baseline}, demonstrate that \tformer outperforms these baselines by a significant margin, with improvements ranging from 3.4\% to 7.5\%.
Interestingly, simple baseline methods like single-frame sampling achieve decent performance, suggesting that the dataset may still exhibit temporal bias. However, \tformer consistently excels, particularly in temporal and causal reasoning, highlighting its superior temporal modeling capabilities.

\begin{table}[tbp]
    \centering
    \setlength{\tabcolsep}{8pt}
    \resizebox{0.8\linewidth}{!}{
    \begin{tabular}{c cccc}
    \toprule
     \multirow{2.5}{*}{\textbf{\#Attn. Heads}}& \multicolumn{4}{c}{\textbf{NExT-QA}} \\
    \cmidrule(lr){2-5}
     & \textbf{Tem.} & \textbf{Cau.} & \textbf{Des.} & \textbf{Avg.} \\
    \midrule    
    2 &  71.5   & 75.8 & 82.6 & 75.5 \\
    4 &  71.8   & 75.9 & 83.7 & 75.8 \\
    8 &  71.9   & 76.5 & 84.1 & 76.2 \\
    12 &  \textbf{72.8}   & \textbf{76.9} & \textbf{84.7} & \textbf{76.7} \\ 
    16 &  72.7   & 76.5 & 84.0 & 76.4 \\ 
    \bottomrule
    \end{tabular}}
    \vspace{-1mm}
    \caption{Effect of the number of attention heads. Increasing the number of attention heads improves the model's performance.} 
    \label{tab:na}
\end{table}

\begin{table}[tbp]
    \centering
    \setlength{\tabcolsep}{8pt}
    \resizebox{0.81\linewidth}{!}{
        \begin{tabular}{ccccc}
        \toprule
         \multirow{2.5}{*}{\textbf{Sampling Ratio}} & \multicolumn{4}{c}{\textbf{NExT-QA}} \\
        \cmidrule(lr){2-5}
         & \textbf{Tem.} & \textbf{Cau.} & \textbf{Des.} & \textbf{Avg.} \\
        \midrule
         $8 \rightarrow 1$ & 70.9 & 75.9 & 82.4 & 74.8 \\
        $16 \rightarrow 1$ & 71.0 & 75.3 & 81.9 & 74.9 \\
         $32 \rightarrow 1$ & 71.4 & 75.4 & 83.2 & 75.3 \\
         $64 \rightarrow 1$ & 71.5 & 75.8 & 82.4 & 75.4 \\
         $128 \rightarrow 1$ & 71.4 & 75.3 & 83.3 & 75.3 \\
         \cellcolor{gray!25} $8 \rightarrow 4$ & \cellcolor{gray!25} 71.9 & \cellcolor{gray!25} 76.8 & \cellcolor{gray!25} \textbf{84.8} & \cellcolor{gray!25} 76.4 \\
         \cellcolor{gray!25} $16 \rightarrow 4$ &\cellcolor{gray!25}  72.8 & \cellcolor{gray!25}76.9 & \cellcolor{gray!25}84.7 & \cellcolor{gray!25}\textbf{76.6} \\
         \cellcolor{gray!25} $32 \rightarrow 4$ & \cellcolor{gray!25}73.0 & \cellcolor{gray!25}76.1 & \cellcolor{gray!25}84.6 &\cellcolor{gray!25} 76.4 \\
         \cellcolor{gray!25} $64 \rightarrow 4$ &\cellcolor{gray!25} 72.4 & \cellcolor{gray!25}\textbf{77.0} & \cellcolor{gray!25}84.3 & \cellcolor{gray!25}\textbf{76.6} \\
         \cellcolor{gray!25} $128 \rightarrow 4$ & \cellcolor{gray!25}\textbf{73.2} & \cellcolor{gray!25}76.6 & \cellcolor{gray!25}84.4 &\cellcolor{gray!25} \textbf{76.6} \\
         $8 \rightarrow 8$ & 71.6 & 75.8 & 83.5 & 75.7 \\
          $16 \rightarrow 8$ & 72.1 & \textbf{77.0} & 84.3 & 76.5 \\
          $32 \rightarrow 8$ & 72.8 & 76.7 & 84.2 & \textbf{76.6} \\
          $64 \rightarrow 8$ & 72.3 & 76.2 & 84.7 & 76.3 \\
        \bottomrule
        \end{tabular}
    }
    \vspace{-1mm}
    \caption{Impact of sampling ratios. We conduct a comprehensive analysis to determine the optimal sampling ratio, defined as ``$\text{\#input frames} \rightarrow \text{\#equivalent output frames}$''.}
    \label{tab:ratio}
\end{table}

\begin{table*}[htbp]
\centering
    \setlength{\tabcolsep}{8pt}
    \resizebox{.95\linewidth}{!}{
         \begin{tabular}{c c c ccccc cccc ccc}
        \toprule
        \multirow{2.5}{*}{\textbf{\tformer}} & \multirow{2.5}{*}{\textbf{Question-Guided}} & \multirow{2.5}{*}{\textbf{Temp. Embed.}}& \multicolumn{4}{c}{\textbf{NExT-QA}}& \multicolumn{5}{c}{\textbf{STAR}}& \multirow{2.5}{*}{\textbf{How2QA}}  & \multirow{2.5}{*}{\textbf{VLEP}} \\
        \cmidrule(lr){4-7} \cmidrule(lr){8-12}
        & & & \textbf{Tem.} & \textbf{Cau.} & \textbf{Des.} & \textbf{Avg.} & \textbf{Int.} & \textbf{Seq.} & \textbf{Pre.} & \textbf{Fea.} & \textbf{Avg.} &&&\\
        \midrule
        \xmark & \xmark & \xmark & 70.3 & 75.8 & 83.4 & 75.4 & 61.0 & 66.3 & 70.3 & 69.8 & 65.1 & 81.2 & 68.6 \\
        \cmark & \xmark & \xmark & 72.2 & 76.0 & 84.6 & 76.1 & 60.8 & 67.2 & 71.6 & 70.4 & 65.4 & 84.4 & 69.0 \\
        \cmark & \cmark & \xmark & 71.5 & 77.0 & 83.9 & 76.3 & 60.8 & 66.8 & 71.2 & 70.2 & 65.4 & 85.6 & 69.3 \\
        \cmark & \xmark & \cmark & \textbf{73.5} & 75.1 & \textbf{85.5} & 76.2 & 63.0 & 68.2 & 70.2 & 72.6 & 66.6 & 84.5 & 69.2 \\
        \cmark & \cmark & \cmark & \underline{72.8} & \textbf{77.0} & \underline {84.7} & \textbf{76.7} & \textbf{62.8} & \textbf{69.6} & \textbf{72.8} & \textbf{70.0} & \textbf{67.6} & \textbf{85.9} & \textbf{69.6} \\
        \bottomrule
        \end{tabular}
    }
    \vspace{-1mm}
\caption{Ablation studies of each module of \framework. Ablating different modules including \tformer, question-guided visual features, and temporal embeddings. The impact on the model's performance demonstrates each module's efficacy.}
\label{tab:ablate}
\vspace{-1mm}
\end{table*}

\begin{figure*}[htbp]
    \centering
    \includegraphics[width=0.95\textwidth]{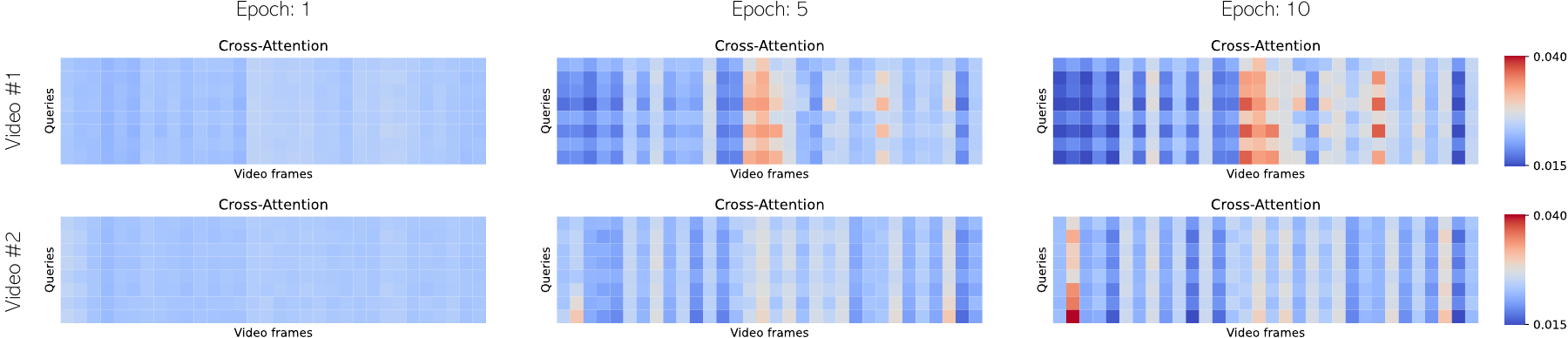}
    \vspace{-1mm}
    \caption{For different video samples (on each row), we visualize the attention map between question-guided temporal queries $\mathbf{q}$ and framewise visual tokens $\mathbf{e}^{1:n}_f$ in the last layer of \tformer during epochs 1, 5, and 10.}
    \label{fig:atmap}
    \vspace{-1mm}
\end{figure*}

\subsection{Exploring Design Choices}\label{sec:design}
We perform a comprehensive analysis of different design choices for \tformer, examining the impact of sampling methods, Transformer settings, and sampling ratios.

\tit{Sampling methods}
We compare four methods for initializing queries $\mathbf{q}_t$: uniform sampling, random sampling, $k$-means sampling, and $k$-medoids sampling. In our experiments, we fix the input frames at 16. 
The clustering-based methods, \ie, $k$-means and $k$-medoids, outperform other techniques, as shown in \cref{tab:sample}. This aligns with our intuition that clustering selects more diverse and crucial frames, providing better initialization.

We also conducted an experiment involving learnable temporal queries akin to the learnable queries in Q-Former. The results, shown in \cref{tab:learnable_ablation}, reveal that our sampling-based approach consistently outperforms learnable temporal queries by a significant margin. This suggests that the initial sampling strategy for temporal queries contributes substantially to the overall performance.

\tit{Investigating \tformer settings}
We investigate the impact of varying our \tformer architecture settings, including the number of hidden layers and attention heads.

In~\cref{tab:nl}, we show that \tformer with two hidden layers represents the most effective solution. We speculate that since Q-Former already captures intricate and abstract visual features, \tformer can easily grasp the temporal modeling of video frames with its lightweight structure.

As shown in~\cref{tab:na}, increasing the number of attention heads improves performance, with 12 heads yielding the best results. We believe that having more attention heads helps to encode diverse relationships between video frames, a critical capability for comprehensive temporal reasoning.

\begin{table*}[htbp]
\centering
\setlength{\tabcolsep}{6pt}
\resizebox{0.98\textwidth}{!}{
    \begin{tabular}{l c c ccccc cccc cc}
    \toprule
    \multirow{2.5}{*}{\textbf{Model}} & \multirow{2.5}{*}{\textbf{w/ LLM}} & \multirow[t]{2}{*}[-0.8mm]{\textbf{\#Params}} & \multicolumn{4}{c}{\textbf{NExT-QA}}& \multicolumn{5}{c}{\textbf{STAR}}& \multirow{2.5}{*}{\textbf{How2QA}} &  \multirow{2.5}{*}{\textbf{VLEP}} \\
    \cmidrule(lr){4-7} \cmidrule(lr){8-12}
    & & \textbf{Train (Tot.)} & \textbf{Tem.} & \textbf{Cau.} & \textbf{Des.} & \textbf{Overall.} & \textbf{Int.} & \textbf{Seq.} & \textbf{Pre.} & \textbf{Fea.} & \textbf{Overall.} &&\\
    \midrule
    Flamingo-9B 4-shot ~\cite{alayrac2022flamingo}  & \cmark & - (9B) & - & - & - & - & - & - & - & - &  42.8 & - & - \\
    Flamingo-9B 32-shot ~\cite{alayrac2022flamingo} & \cmark & - (9B) & - & - & - & - & - & - & - & - &  41.2 & -  & - \\
    Flamingo-80B 4-shot ~\cite{alayrac2022flamingo} & \cmark & - (80B) & - & - & - & - & - & - & - & - &  42.4 & - & - \\
    Flamingo-80B 32-shot ~\cite{alayrac2022flamingo} & \cmark & - (80B) & - & - & - & - & - & - & - & - &  42.2 & - & - \\
    FrozenBiLM ~\cite{yang2022zero} & \xmark & 30M (30M) & -& -& - & - & - & - & - & - & - & 81.5  & - \\
    All-in-One ~\cite{wang2023all} &  \xmark & 110M (110M) & 48.6 & 48.0 & 63.2 & 50.6 & 47.5 & 50.8 & 47.7 & 44.0 & 47.5 & - & - \\
    ATP ~\cite{buch2022revisiting}& \xmark& - & 53.1 & 50.2 & 66.8 & 54.3 & 50.6 & 52.9 & 49.4&  40.6& 48.4& -& -  \\
    MIST ~\cite{gao2023mist} & \xmark & - & 56.6 & 54.6 & 66.9 & 57.1 & 55.5 & 54.2 & 54.2 & 44.4 & 51.1 & - & - \\
    TranSTR ~\cite{li2023discovering} & \xmark &-  & 59.7 & 60.2 & 70.0 & 61.5 &- & - & - & - & - & -  & - \\
    HiTeA~\cite{ye2023hitea} & \xmark & - & 58.3 & 62.4 & 75.6 & 63.1 & - & - & - & - & - & -  & - \\
    InternVideo~\cite{wang2022internvideo} & \xmark & 1.3B (1.3B) & 58.5  & 62.5 & 75.8 & 63.2 & 62.7 & 65.6 & 54.9 & 51.9 & 58.7 & 79.0  & 63.9 \\
    BLIP-2$^{\text{voting}}$ ~\cite{li2023blip} & \cmark & 188M (7.8B) & 65.2 & 70.1 & 80.1 & 70.1 & 52.3 & 54.8 & 49.0& 51.2 & 51.8 & 79.6  & 67.0 \\ 
    BLIP-2$^{\text{concat}}$ ~\cite{li2023blip}  & \cmark & 188M (7.8B) & 68.1  & 72.9 & 81.2 & 72.6 & \underline{65.4} & 69.0 & 59.7 & 54.2  & 62.0 & 82.2  & 68.6 \\
    SeViLa ~\cite{yu2023self} & \cmark & 106M (4.1B) & 69.4 & \underline{74.2} & \underline{81.3} & \underline{73.8} & 63.7 & \textbf{70.4} & \underline{63.1} & \underline{62.4} & 64.9 & \underline{83.6} & 68.9 \\
    LLaMA-VQA \cite{ko2023large} & \cmark  & 4.5M (7B) & \underline{72.7}  &69.2 & 75.8 & 72.0 & \textbf{66.2} & 67.9 & 57.2 & 52.7 &  \underline{65.4} & -  & \textbf{71.0} \\
    \midrule
    \textbf{\framework (Ours)} & \cmark & 50.1M (7.8B) & \textbf{72.8} & \textbf{76.9} & \textbf{84.7} & \textbf{76.7 (\textcolor{green}{+2.9})} & 62.8 & \underline{69.6} & \textbf{72.8} & \textbf{70.0} & \textbf{67.6 (\textcolor{green}{+2.2})} & \textbf{85.9 (\textcolor{green}{+2.3})}  & \underline{69.6} \\
    \bottomrule
    \end{tabular}
}
\caption{Comparison with SOTA video QA models. NExT-QA
involves \textit{tem}poral, \textit{cau}sal, and \textit{des}criptive question types. STAR contains four question types: \textit{int}eraction, \textit{seq}uence,
\textit{pre}diction, and \textit{fea}sibility. We emphasize the best performance in \textbf{bold}, and \underline{underline} the second-best performance. \framework consistently outperforms SOTA models. 
Particularly, it manifests great temporal and causal reasoning capabilities on NeXT-QA and STAR. These results underscore the effectiveness of \framework's temporal modeling for VideoQA tasks.
}
\label{tab:main_ft}
\end{table*}

\tit{Impact of video-frame sampling ratio}
We analyze how the sampling ratio, defined as ``$\text{number of input frames} \rightarrow \text{numbers of equivalent output frames}$'', influences the performance of \ours.
A higher sampling ratio compresses the input frame sequence into a smaller number of output tokens, referred to as the number of equivalent output frames. Conversely, a lower sampling ratio produces more fine-grained representations, but it also leads to longer input sequences for LLMs.

As shown in \cref{tab:ratio}, for a fixed sampling ratio, such as $8\rightarrow1$ and $32\rightarrow4$, larger output sizes consistently yield better performance by accommodating more information. Conversely, compressing entire video frames into a small number of tokens while keeping the input frame size constant, as in $16\rightarrow1$ and $16\rightarrow4$, can hinder performance.
Moreover, when the output frame size is fixed, as in $16\rightarrow4$, $32\rightarrow4$, and $64\rightarrow4$, incorporating more input frames enhances the performance by introducing additional information. However, there is an upper limit beyond which adding more frames no longer enhances performance.

Our findings confirm that increasing the number of input frames improves the performance, but only up to a certain limit. While additional frames provide more question-aware information, they may also introduce excessive redundancy. Additionally, the number of output frames is critical for achieving optimal results. We found that using four output frames consistently yields the best results, irrespective of the number of input frames.

We hypothesize that an insufficient number of output frames may fail to capture enough information, while a large number could introduce redundancy. Longer sequences may also divert the attention of LLMs, potentially compromising performance.
Furthermore, the absolute number of output frames outweighs a fixed sampling ratio, highlighting the importance of carefully selecting the number of output frames to ensure optimal performance.

\subsection{Ablation Study}
To assess the contribution of various components in \ours, we conduct an ablation study, as detailed in \cref{tab:ablate}. We ablate \tformer, question-guided queries, and temporal embeddings. Our results indicate that all key components contribute to improved performance across multiple datasets. Notably, the absence of \tformer results in significant performance degradation, highlighting its critical role. Additionally, leveraging question tokens to extract question-guided visual features also enhances performance. This collective improvement underscores the success and efficacy of our question-guided, localization-free spatio-temporal feature extraction approach.

\subsection{Visualizing \tformer Attention Mechanisms}
To highlight the learning behavior of \tformer, we attempt to visualize the query-video cross-attention maps. \cref{fig:atmap} showcases the attention map between question-guided temporal queries $\mathbf{q}$ and framewise visual tokens $\mathbf{e}^{1:n}_f$ in the last layer of \tformer across epochs 1, 5, and 10.

We observe that throughout the training, \tformer effectively learns the temporal dependencies between frames at different timestamps across the entire video sequence. The frames with the highest attention scores are distributed across the video, indicating \tformer's ability to discern the importance of different frames. This also demonstrates that \tformer can attend to and differentiate the significance of frames, enabling it to identify the most crucial frames without relying on explicit localization methods.

\subsection{Comparison to State-of-the-art Models}
Finally, we compare our \ours with state-of-the-art models across all four benchmarks, as presented in \cref{tab:main_ft}. Notably, \ours consistently outperforms existing models.

On NExT-QA, our model achieves the best performance in the overall category, with a 2.9\% improvement. It also excels in the three subcategories that emphasize temporal modeling and causal reasoning, demonstrating the effectiveness of \tformer. Similarly, on the STAR dataset, \framework achieves a 2.2\% performance boost over the current benchmark and attains the best performance in each subcategory. On How2QA, our model surpasses all others by 2.3\%.

Our results reveal a significant performance boost compared to non-LLM baseline models, confirming the potency of LLMs as robust visual reasoners when provided with appropriate visual context. Moreover, \framework surpasses LLM-based competitors, particularly excelling in causal reasoning on NExT-QA and event prediction on STAR, further validating the efficacy of \ours.

Notably, \ours utilizes a smaller LLM as the reasoning backbone and a limited number of learnable parameters compared to other models. Furthermore, unlike most competitors that rely on extensive video-based pre-training, our model is trained exclusively on the target dataset. This highlights the efficiency and effectiveness of our approach.

%% file: sections/5-Conclusion.tex
\section{Conclusion}
In this work, we introduced \tformer, a novel question-guided temporal querying Transformer designed to address the challenges of dense video processing for multiple-choice video QA tasks. Our integrated framework, named \ours, combines \tformer with LLMs as reasoning agents, significantly outperforming other temporal modeling methods. Our approach also achieves substantial improvements over existing LLM-based models across four challenging video QA benchmarks while being trained solely on the target datasets without requiring any video-based pre-training phase.
Through extensive experimental evaluations, we showcased the effectiveness of our approach. By leveraging compact, question-guided temporal tokens, we successfully bridge the gap between the complex temporal dynamics in videos and the reasoning capabilities of LLMs.

In future studies, we aim to extend the application of \tformer to a broader spectrum of video understanding tasks, including free-form video QA and more general video reasoning tasks.
We believe that \tformer holds significant potential as a robust adaptation mechanism for MLLMs across diverse video understanding applications.

%% file: sections/X-Suppl.tex
\section*{Appendix}
This appendix provides additional implementation details about our \framework model. The sections delve into critical aspects, including ablation studies, hyperparameter configurations, 
and a thorough analysis of the impact of linguistic bias on performance.

\section{Ablation Studies}
In this section, we present additional ablation studies to further analyze our model's behavior. Section \ref{sec:bottle} investigates the effect of varying the number and dimensionality of the layers in \tformer. Section \ref{sec:ling_bias} explores the extent of linguistic bias within benchmark datasets.

\subsection{\tformer settings}\label{sec:bottle}
We investigate the impact of layer number and intermediate dimensionality in the feed-forward layer of the \tformer, as shown in Tab.~\ref{tab:bottle}. Our experiments demonstrate that increasing the number of hidden layers improves model performance, while larger bottleneck dimensionality yields the opposite effect. Our findings suggest that a configuration of 2 hidden layers with a 768-dimensional feed-forward layer yields the best performance.

\subsection{Exploring Linguistic Bias}\label{sec:ling_bias}
Linguistic bias in video question-answering datasets is a significant concern when using Large Language Models (LLMs). We conduct a comprehensive analysis to verify if the questions in the benchmarks contain biases that enable models to answer correctly without visual input.

\begin{figure*}[!t]
    \centering
    \begin{subfigure}[b]{0.45\textwidth}
        \centering
        \includegraphics[width=\textwidth]{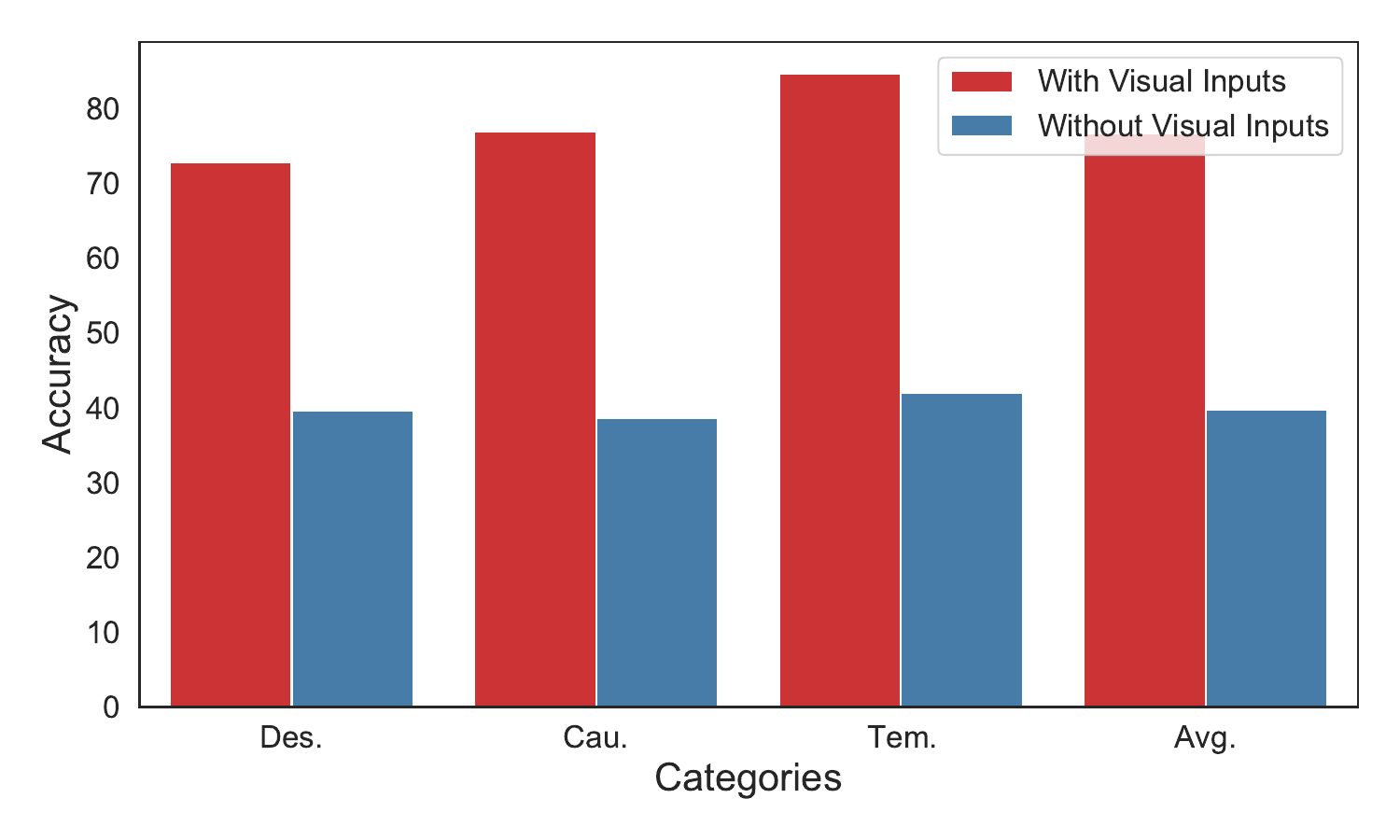}
        \caption{NextQA}
        \label{subfig:next}
    \end{subfigure}
    \begin{subfigure}[b]{0.45\textwidth}
        \centering
        \includegraphics[width=\textwidth]{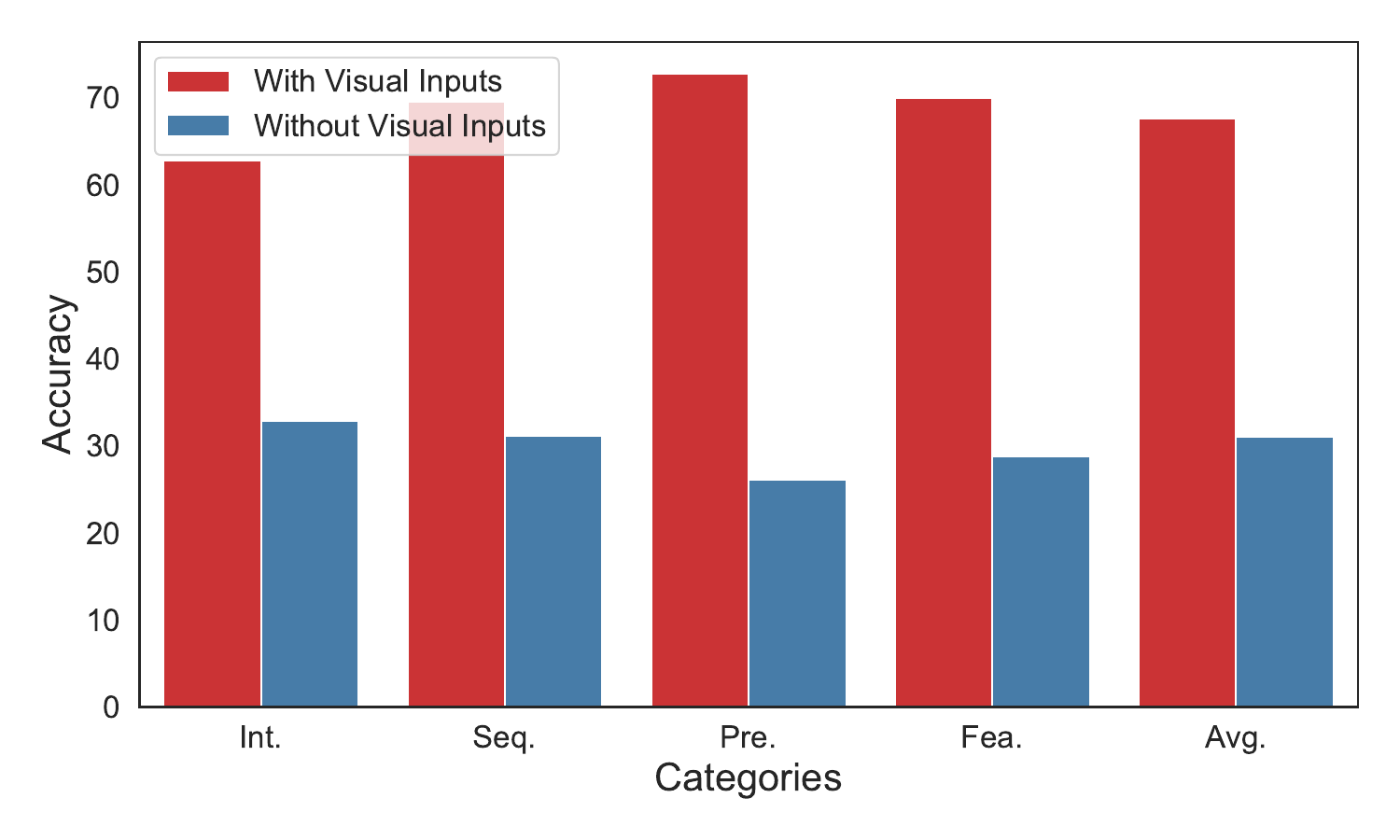}
        \caption{STAR}
        \label{subfig:star}
    \end{subfigure}
    \caption{Exploring Linguistic Bias. We observe that LLM reasoners can only achieve a modest performance, akin to a ``blind guess'' when visual inputs are absent.}
    \label{fig:linguisti}
    \vspace{15cm}
\end{figure*}

Fig.~\ref{fig:linguisti} presents the full performance results across different datasets and categories. Our observations indicate that in the absence of visual information, LLM reasoners exhibit modest performance, comparable to a ``blind guess''. This finding highlights the robustness of the video question-answering benchmarks, ensuring they are minimally influenced by linguistic bias.

Notably, the performance gap between our model and LLM reasoners is even more pronounced in categories such as causal and temporal reasoning. This underscores the effectiveness of our approach in leveraging visual information rather than being overly dependent on linguistic cues.

\begin{table}[t]
\centering
\small
\setlength{\tabcolsep}{6pt}
\resizebox{0.85\linewidth}{!}{
\begin{tabular}{cc cccc}
\toprule
\multirow{2.5}{*}{\shortstack{\\ \textbf{\#Linear} \\ \textbf{Layers}}} & \multirow{2.5}{*}{\shortstack{\\ \textbf{Bottleneck} \\ \textbf{Dim}}} & \multicolumn{4}{c}{\textbf{NExT-QA}}  \\
\cmidrule(lr){3-6}
 & & \textbf{Tem.} & \textbf{Cau.} & \textbf{Des.} & \textbf{Avg.} \\
\midrule
2 & 3,072  & 72.4   & 75.8 & 81.7 & 75.7 \\
2 & 1,536  & 72.5   & \textbf{77.0} & 82.5 & 76.4 \\
2 & 768  & \textbf{72.8}   & 76.9 & \textbf{84.7} & \textbf{76.7} \\
1 & 768  & 71.7   & 75.0 & 82.9 & 75.2 \\
\bottomrule
\end{tabular}}
\caption{Effect of different feed-forward bottleneck size. Increasing the number of linear layers improves the model performance, but larger bottleneck dimensionality affects the results.
}
\label{tab:bottle}
\end{table}

\begin{table}[t]
\centering
\scriptsize
\setlength{\tabcolsep}{7pt}
\resizebox{1.\linewidth}{!}{
\begin{tabular}{l c  cc  c}
\toprule
\textbf{\multirow{2.5}{*}{\shortstack{\\ Method }} } &  \multirow{2.5}{*}{\shortstack{\\ \textbf{\#Pre-train} \\ \textbf{videos/images}}} & \multicolumn{2}{c}{\textbf{TGIF}} & \multirow{2.5}{*}{\shortstack{\\ \textbf{MSRVTT} \\ \textbf{MC}}} \\
\cmidrule(lr){3-4}
&  & \textbf{Act.} & \textbf{Trans.} &  \\
\midrule
All-in-one & 283M & 95.5 & 94.7 & 92.3 \\
VIOLET & 186M & 92.5 & 95.7 & 91.9 \\
MERLOT & 180M & 94.0 & 96.2 & 90.2  \\
Singularity & 17M & - & - & 92.1 \\
Clover & 5M & 94.9 & 98.0 & 95.0 \\
ClipBERT & 200k & 82.8 & 87.8 & 88.2 \\
\midrule
\textbf{PQR (Ours)} & \textbf{0} & 96.1 & 98.4 & 96.2 \\
\bottomrule
\end{tabular}}
\caption{Extending comparison to additional datasets. Our \framework consistently outperforms other baseline models despite being trained solely on the target dataset.
}
\label{tab:new_ds}
\end{table}

\subsection{Extended Results}
We further evaluate \framework on the TGIF-QA~\cite{jang2017tgif} and MSRVTT-MC~\cite{xu2016msr} datasets, as shown in Tab.~\ref{tab:new_ds}. Notably, \framework is trained solely on the target dataset without requiring extensive pre-training on millions of videos. Despite this, it consistently outperforms other baseline models.

\section{Additional Implementation Details}

\begin{table}[!t]
\centering
\scriptsize
\setlength{\tabcolsep}{4pt}
\resizebox{1\linewidth}{!}{
 \begin{tabular}{l c c c c c}
\toprule
\textbf{Parameter} & \textbf{NExT-QA} & \textbf{STAR} & \textbf{How2QA} & \textbf{VLEP} \\
\midrule
Batch Size & 2 & 2 & 4 & 4 \\
Epochs & 10 & 10 & 10 & 10 \\
Iterations per Epoch & 2500 & 5000 & 5000 & 1000 \\
Warmup Epochs & 1 & 1 & 1 & 1 \\
Cooldown Epochs & 2 & 2 & 5 & 5 \\
Initial LR & 3e-5 & 5e-5 & 3e-5 & 2e-5 \\
Warmup LR & 8e-6 & 1e-5 & 8e-6 & 7e-6 \\
Minimum LR & 1e-6 & 1e-6 & 1e-6 & 1e-6 \\
\bottomrule
\end{tabular}}
\caption{\framework training hyperparameters for different datasets.}
\label{tab:hyperparameters}
\end{table}

\subsection{Hyperparameters}
In this section, we provide a detailed overview of the training hyperparameters used across all benchmark datasets to unsure reproducibility. Tab.~\ref{tab:hyperparameters} presents the optimal values for key parameters, including batch size, total epoch numbers, number of iteration steps per epoch, warm-up and cooldown epochs, and learning rate.

%% file: main.bbl
\begin{thebibliography}{10}\itemsep=-1pt

\bibitem{alayrac2022flamingo}
Jean-Baptiste Alayrac, Jeff Donahue, Pauline Luc, Antoine Miech, Iain Barr, Yana Hasson, Karel Lenc, Arthur Mensch, Katherine Millican, Malcolm Reynolds, et~al.
\newblock Flamingo: a visual language model for few-shot learning.
\newblock {\em NeurIPS}, 2022.

\bibitem{bain2021frozen}
Max Bain, Arsha Nagrani, G{\"u}l Varol, and Andrew Zisserman.
\newblock Frozen in time: A joint video and image encoder for end-to-end retrieval.
\newblock In {\em ICCV}, 2021.

\bibitem{buch2022revisiting}
Shyamal Buch, Crist{\'o}bal Eyzaguirre, Adrien Gaidon, Jiajun Wu, Li Fei-Fei, and Juan~Carlos Niebles.
\newblock Revisiting the" video" in video-language understanding.
\newblock In {\em CVPR}, 2022.

\bibitem{caffagni2024r}
Davide Caffagni, Federico Cocchi, Luca Barsellotti, Nicholas Moratelli, Sara Sarto, Lorenzo Baraldi, Marcella Cornia, and Rita Cucchiara.
\newblock The (r) evolution of multimodal large language models: A survey.
\newblock {\em ACL}, 2024.

\bibitem{chen2022pali}
Xi Chen, Xiao Wang, Soravit Changpinyo, AJ Piergiovanni, Piotr Padlewski, Daniel Salz, Sebastian Goodman, Adam Grycner, Basil Mustafa, Lucas Beyer, et~al.
\newblock Pali: A jointly-scaled multilingual language-image model.
\newblock {\em ICLR}, 2023.

\bibitem{chung2022scaling}
Hyung~Won Chung, Le Hou, Shayne Longpre, Barret Zoph, Yi Tay, William Fedus, Yunxuan Li, Xuezhi Wang, Mostafa Dehghani, Siddhartha Brahma, et~al.
\newblock Scaling instruction-finetuned language models.
\newblock {\em JMLR}, 2024.

\bibitem{dai2023instructblip}
Wenliang Dai, Junnan Li, Dongxu Li, Anthony Meng~Huat Tiong, Junqi Zhao, Weisheng Wang, Boyang Li, Pascale Fung, and Steven Hoi.
\newblock Instructblip: Towards general-purpose vision-language models with instruction tuning.
\newblock In {\em NIPS}, 2023.

\bibitem{fang2023eva}
Yuxin Fang, Wen Wang, Binhui Xie, Quan Sun, Ledell Wu, Xinggang Wang, Tiejun Huang, Xinlong Wang, and Yue Cao.
\newblock Eva: Exploring the limits of masked visual representation learning at scale.
\newblock In {\em CVPR}, 2023.

\bibitem{gao2023mist}
Difei Gao, Luowei Zhou, Lei Ji, Linchao Zhu, Yi Yang, and Mike~Zheng Shou.
\newblock Mist: Multi-modal iterative spatial-temporal transformer for long-form video question answering.
\newblock In {\em CVPR}, 2023.

\bibitem{grunde2021agqa}
Madeleine Grunde-McLaughlin, Ranjay Krishna, and Maneesh Agrawala.
\newblock Agqa: A benchmark for compositional spatio-temporal reasoning.
\newblock In {\em CVPR}, 2021.

\bibitem{huang2021tada}
Ziyuan Huang, Shiwei Zhang, Liang Pan, Zhiwu Qing, Mingqian Tang, Ziwei Liu, and Marcelo~H Ang~Jr.
\newblock Tada! temporally-adaptive convolutions for video understanding.
\newblock {\em ICLR}, 2022.

\bibitem{jang2017tgif}
Yunseok Jang, Yale Song, Youngjae Yu, Youngjin Kim, and Gunhee Kim.
\newblock Tgif-qa: Toward spatio-temporal reasoning in visual question answering.
\newblock In {\em CVPR}, 2017.

\bibitem{ko2023large}
Dohwan Ko, Ji~Soo Lee, Wooyoung Kang, Byungseok Roh, and Hyunwoo~J Kim.
\newblock Large language models are temporal and causal reasoners for video question answering.
\newblock In {\em EMNLP}, 2023.

\bibitem{kojima2022large}
Takeshi Kojima, Shixiang~Shane Gu, Machel Reid, Yutaka Matsuo, and Yusuke Iwasawa.
\newblock Large language models are zero-shot reasoners.
\newblock {\em NeurIPS}, 2022.

\bibitem{le2020neural}
Thao~Minh Le, Vuong Le, Svetha Venkatesh, and Truyen Tran.
\newblock Neural reasoning, fast and slow, for video question answering.
\newblock In {\em IJCNN}, 2020.

\bibitem{lei2022revealing}
Jie Lei, Tamara~L Berg, and Mohit Bansal.
\newblock Revealing single frame bias for video-and-language learning.
\newblock {\em ACL}, 2023.

\bibitem{lei2021less}
Jie Lei, Linjie Li, Luowei Zhou, Zhe Gan, Tamara~L Berg, Mohit Bansal, and Jingjing Liu.
\newblock Less is more: Clipbert for video-and-language learning via sparse sampling.
\newblock In {\em CVPR}, 2021.

\bibitem{lei2018tvqa}
Jie Lei, Licheng Yu, Mohit Bansal, and Tamara~L Berg.
\newblock Tvqa: Localized, compositional video question answering.
\newblock In {\em EMNLP}, 2018.

\bibitem{lei2020more}
Jie Lei, Licheng Yu, Tamara Berg, and Mohit Bansal.
\newblock What is more likely to happen next? video-and-language future event prediction.
\newblock In {\em EMNLP}, 2020.

\bibitem{li2023blip}
Junnan Li, Dongxu Li, Silvio Savarese, and Steven Hoi.
\newblock Blip-2: Bootstrapping language-image pre-training with frozen image encoders and large language models.
\newblock {\em ICML}, 2023.

\bibitem{li2020hero}
Linjie Li, Yen-Chun Chen, Yu Cheng, Zhe Gan, Licheng Yu, and Jingjing Liu.
\newblock Hero: Hierarchical encoder for video+ language omni-representation pre-training.
\newblock In {\em EMNLP}, 2020.

\bibitem{li2019beyond}
Xiangpeng Li, Jingkuan Song, Lianli Gao, Xianglong Liu, Wenbing Huang, Xiangnan He, and Chuang Gan.
\newblock Beyond rnns: Positional self-attention with co-attention for video question answering.
\newblock In {\em AAAI}, 2019.

\bibitem{li2023discovering}
Yicong Li, Junbin Xiao, Chun Feng, Xiang Wang, and Tat-Seng Chua.
\newblock Discovering spatio-temporal rationales for video question answering.
\newblock In {\em ICCV}, 2023.

\bibitem{liu2024llavanext}
Haotian Liu, Chunyuan Li, Yuheng Li, Bo Li, Yuanhan Zhang, Sheng Shen, and Yong~Jae Lee.
\newblock Llava-next: Improved reasoning, ocr, and world knowledge, 2024.

\bibitem{liu2023llava}
Haotian Liu, Chunyuan Li, Qingyang Wu, and Yong~Jae Lee.
\newblock Visual instruction tuning, 2023.

\bibitem{liu2023evaluating}
Hanmeng Liu, Ruoxi Ning, Zhiyang Teng, Jian Liu, Qiji Zhou, and Yue Zhang.
\newblock Evaluating the logical reasoning ability of chatgpt and gpt-4.
\newblock {\em arXiv preprint arXiv:2304.03439}, 2023.

\bibitem{luo2021clip4clip}
Huaishao Luo, Lei Ji, Ming Zhong, Yang Chen, Wen Lei, Nan Duan, and Tianrui Li.
\newblock Clip4clip: An empirical study of clip for end to end video clip retrieval.
\newblock {\em Neurocomputing}, 2022.

\bibitem{Maaz2023VideoChatGPT}
Muhammad Maaz, Hanoona Rasheed, Salman Khan, and Fahad~Shahbaz Khan.
\newblock Video-chatgpt: Towards detailed video understanding via large vision and language models.
\newblock In {\em ACL}, 2024.

\bibitem{peng2021progressive}
Liang Peng, Shuangji Yang, Yi Bin, and Guoqing Wang.
\newblock Progressive graph attention network for video question answering.
\newblock In {\em ACM Multimedia}, 2021.

\bibitem{radford2021learning}
Alec Radford, Jong~Wook Kim, Chris Hallacy, Aditya Ramesh, Gabriel Goh, Sandhini Agarwal, Girish Sastry, Amanda Askell, Pamela Mishkin, Jack Clark, et~al.
\newblock Learning transferable visual models from natural language supervision.
\newblock In {\em ICML}, 2021.

\bibitem{tang2023video}
Yunlong Tang, Jing Bi, Siting Xu, Luchuan Song, Susan Liang, Teng Wang, Daoan Zhang, Jie An, Jingyang Lin, Rongyi Zhu, Ali Vosoughi, Chao Huang, Zeliang Zhang, Feng Zheng, Jianguo Zhang, Ping Luo, Jiebo Luo, and Chenliang Xu.
\newblock Video understanding with large language models: A survey.
\newblock {\em arXiv preprint arXiv:2312.17432}, 2023.

\bibitem{alpaca}
Rohan Taori, Ishaan Gulrajani, Tianyi Zhang, Yann Dubois, Xuechen Li, Carlos Guestrin, Percy Liang, and Tatsunori~B. Hashimoto.
\newblock Stanford alpaca: An instruction-following llama model, 2023.

\bibitem{tapaswi2016movieqa}
Makarand Tapaswi, Yukun Zhu, Rainer Stiefelhagen, Antonio Torralba, Raquel Urtasun, and Sanja Fidler.
\newblock Movieqa: Understanding stories in movies through question-answering.
\newblock In {\em CVPR}, 2016.

\bibitem{touvron2023llama}
Hugo Touvron, Thibaut Lavril, Gautier Izacard, Xavier Martinet, Marie-Anne Lachaux, Timoth{\'e}e Lacroix, Baptiste Rozi{\`e}re, Naman Goyal, Eric Hambro, Faisal Azhar, et~al.
\newblock Llama: Open and efficient foundation language models.
\newblock {\em arXiv preprint arXiv:2302.13971}, 2023.

\bibitem{vaswani2017attention}
Ashish Vaswani, Noam Shazeer, Niki Parmar, Jakob Uszkoreit, Llion Jones, Aidan~N Gomez, {\L}ukasz Kaiser, and Illia Polosukhin.
\newblock Attention is all you need.
\newblock {\em NeurIPS}, 2017.

\bibitem{wang2023all}
Jinpeng Wang, Yixiao Ge, Rui Yan, Yuying Ge, Kevin~Qinghong Lin, Satoshi Tsutsui, Xudong Lin, Guanyu Cai, Jianping Wu, Ying Shan, et~al.
\newblock All in one: Exploring unified video-language pre-training.
\newblock In {\em CVPR}, 2023.

\bibitem{wang2022internvideo}
Yi Wang, Kunchang Li, Yizhuo Li, Yinan He, Bingkun Huang, Zhiyu Zhao, Hongjie Zhang, Jilan Xu, Yi Liu, Zun Wang, et~al.
\newblock Internvideo: General video foundation models via generative and discriminative learning.
\newblock {\em arXiv preprint arXiv:2212.03191}, 2022.

\bibitem{wu2021star}
Bo Wu, Shoubin Yu, Tenenbaum Joshua~B Chen, Zhenfang, and Chuang Gan.
\newblock Star: A benchmark for situated reasoning in real-world videos.
\newblock In {\em NeurIPS}, 2021.

\bibitem{xiao2021next}
Junbin Xiao, Xindi Shang, Angela Yao, and Tat-Seng Chua.
\newblock Next-qa: Next phase of question-answering to explaining temporal actions.
\newblock In {\em CVPR}, 2021.

\bibitem{xiao2022video}
Junbin Xiao, Pan Zhou, Tat-Seng Chua, and Shuicheng Yan.
\newblock Video graph transformer for video question answering.
\newblock In {\em ECCV}, 2022.

\bibitem{xu2017video}
Dejing Xu, Zhou Zhao, Jun Xiao, Fei Wu, Hanwang Zhang, Xiangnan He, and Yueting Zhuang.
\newblock Video question answering via gradually refined attention over appearance and motion.
\newblock In {\em ACM Multimedia}, 2017.

\bibitem{xu2016msr}
Jun Xu, Tao Mei, Ting Yao, and Yong Rui.
\newblock Msr-vtt: A large video description dataset for bridging video and language.
\newblock In {\em CVPR}, 2016.

\bibitem{xu2023multimodal}
Peng Xu, Xiatian Zhu, and David~A Clifton.
\newblock Multimodal learning with transformers: A survey.
\newblock {\em TPAMI}, 2023.

\bibitem{yang2021just}
Antoine Yang, Antoine Miech, Josef Sivic, Ivan Laptev, and Cordelia Schmid.
\newblock Just ask: Learning to answer questions from millions of narrated videos.
\newblock In {\em ICCV}, 2021.

\bibitem{yang2022zero}
Antoine Yang, Antoine Miech, Josef Sivic, Ivan Laptev, and Cordelia Schmid.
\newblock Zero-shot video question answering via frozen bidirectional language models.
\newblock {\em NeurIPS}, 2022.

\bibitem{yao2022react}
Shunyu Yao, Jeffrey Zhao, Dian Yu, Nan Du, Izhak Shafran, Karthik Narasimhan, and Yuan Cao.
\newblock React: Synergizing reasoning and acting in language models.
\newblock {\em ICLR}, 2023.

\bibitem{ye2023hitea}
Qinghao Ye, Guohai Xu, Ming Yan, Haiyang Xu, Qi Qian, Ji Zhang, and Fei Huang.
\newblock Hitea: Hierarchical temporal-aware video-language pre-training.
\newblock In {\em ICCV}, 2023.

\bibitem{yu2023self}
Shoubin Yu, Jaemin Cho, Prateek Yadav, and Mohit Bansal.
\newblock Self-chained image-language model for video localization and question answering.
\newblock {\em NeurIPS}, 2024.

\bibitem{yu2024connecting}
Wenyi Yu, Changli Tang, Guangzhi Sun, Xianzhao Chen, Tian Tan, Wei Li, Lu Lu, Zejun Ma, and Chao Zhang.
\newblock Connecting speech encoder and large language model for asr.
\newblock In {\em ICASSP}, 2024.

\bibitem{yu2019activitynet}
Zhou Yu, Dejing Xu, Jun Yu, Ting Yu, Zhou Zhao, Yueting Zhuang, and Dacheng Tao.
\newblock Activitynet-qa: A dataset for understanding complex web videos via question answering.
\newblock In {\em AAAI}, 2019.

\bibitem{zhang2023multi}
Gengyuan Zhang, Jisen Ren, Jindong Gu, and Volker Tresp.
\newblock Multi-event video-text retrieval.
\newblock In {\em ICCV}, 2023.

\bibitem{damonlpsg2023videollama}
Hang Zhang, Xin Li, and Lidong Bing.
\newblock Video-llama: An instruction-tuned audio-visual language model for video understanding.
\newblock {\em EMNLP}, 2023.

\bibitem{zhao2023mmicl}
Haozhe Zhao, Zefan Cai, Shuzheng Si, Xiaojian Ma, Kaikai An, Liang Chen, Zixuan Liu, Sheng Wang, Wenjuan Han, and Baobao Chang.
\newblock Mmicl: Empowering vision-language model with multi-modal in-context learning.
\newblock {\em ICLR}, 2024.

\bibitem{zheng2023judging}
Lianmin Zheng, Wei-Lin Chiang, Ying Sheng, Siyuan Zhuang, Zhanghao Wu, Yonghao Zhuang, Zi Lin, Zhuohan Li, Dacheng Li, Eric Xing, et~al.
\newblock Judging llm-as-a-judge with mt-bench and chatbot arena.
\newblock {\em NIPS}, 2023.

\bibitem{zhong2022video}
Yaoyao Zhong, Junbin Xiao, Wei Ji, Yicong Li, Weihong Deng, and Tat-Seng Chua.
\newblock Video question answering: Datasets, algorithms and challenges.
\newblock {\em EMNLP}, 2022.

\bibitem{zhou2017adaptive}
Yizhou Zhou, Xiaoyan Sun, Dong Liu, Zhengjun Zha, and Wenjun Zeng.
\newblock Adaptive pooling in multi-instance learning for web video annotation.
\newblock In {\em ICCVW}, 2017.

\bibitem{zhu2023minigpt}
Deyao Zhu, Jun Chen, Xiaoqian Shen, Xiang Li, and Mohamed Elhoseiny.
\newblock Minigpt-4: Enhancing vision-language understanding with advanced large language models.
\newblock {\em ICLR}, 2024.

\end{thebibliography}
